\pdfoutput=1

\documentclass[11pt]{article}

\usepackage[]{acl}

\usepackage{times}
\usepackage{latexsym}
\usepackage{booktabs}
\usepackage{tikz}
\usepackage{xcolor}
\usepackage{enumitem}
\usepackage{fontawesome5}
\usepackage[utf8]{inputenc}   
\usepackage{csquotes}         
\usepackage{graphicx}
\usepackage{subcaption}
\usepackage{tcolorbox}
\usepackage{amsmath,amssymb}
\usepackage{tabularx}
\usepackage{geometry}
\usepackage{longtable}
\usepackage{array}
\usepackage{ltablex} 

\usepackage[T1]{fontenc}

\usepackage[utf8]{inputenc}

\usepackage{microtype}

\usepackage{inconsolata}

\usepackage{graphicx}

\title{When AIs Judge AIs: The Rise of Agent-as-a-Judge Evaluation for LLMs}
\author{Fangyi Yu}

\begin{document}

\maketitle

\begin{abstract}
As large language models (LLMs) grow in capability and autonomy, evaluating their outputs—especially in open-ended and complex tasks—has become a critical bottleneck. A new paradigm is emerging: using AI agents as the evaluators themselves. This “agent-as-a-judge” approach leverages the reasoning and perspective-taking abilities of LLMs to assess the quality and safety of other models, promising scalable and nuanced alternatives to human evaluation. In this review, we define the agent-as-a-judge concept, trace its evolution from single-model judges to dynamic multi-agent debate frameworks, and critically examine their strengths and shortcomings. We compare these approaches across reliability, cost, and human alignment, and survey real-world deployments in domains such as medicine, law, finance, and education. Finally, we highlight pressing challenges—including bias, robustness, and meta-evaluation—and outline future research directions. By bringing together these strands, our review demonstrates how agent-based judging can complement (but not replace) human oversight, marking a step toward trustworthy, scalable evaluation for next-generation LLMs.
\end{abstract}

\section{Introduction}
Evaluating the performance of large language models and AI agents is a critical but challenging task. Traditional evaluation methods rely on either human judgment or automatic metrics, each with shortcomings. Human evaluations are the gold standard for subjective criteria like usefulness or coherence, but they are costly, time-consuming, and hard to scale \cite{Mohammadi_2025}. Automated metrics based on n-gram overlap (e.g., BLEU \cite{10.3115/1073083.1073135}, ROUGE \cite{lin-2004-rouge}) provide quick quantitative feedback but often correlate poorly with human judgment on open-ended tasks \cite{novikova-etal-2017-need}. As a result, researchers have turned to the models themselves as potential evaluators, giving rise to the LLM-as-a-judge paradigm \cite{zheng2023judgingllmasajudgemtbenchchatbot}. In this approach, a strong LLM (such as GPT-4.1) is prompted to rate or rank the outputs of other models based on quality, correctness, and other criteria, thereby simulating a human evaluator. Prior studies have shown that a well-prompted LLM can mimic human evaluation reasonably well, achieving scores aligned with human preferences in dialogue and summarization tasks \cite{zheng2023judgingllmasajudgemtbenchchatbot}. This makes LLM-as-a-judge a scalable and transparent alternative to manual annotation.

However, early LLM-as-judge methods typically involve a single model passing judgment, which introduces its own limitations. A single LLM judge may carry inherent biases – for instance, preferring certain writing styles or content – and thus produce skewed evaluations \cite{koo-etal-2024-benchmarking}. It also represents only one perspective, whereas complex real-world evaluations often require balancing multiple criteria or viewpoints \cite{he-etal-2023-medeval, chen2025multiagentasjudgealigningllmagentbasedautomated}. Human evaluation processes usually involve multiple annotators to mitigate individual bias and inconsistency. Recognizing this, researchers began exploring multi-agent evaluation frameworks in which multiple LLM agents collaborate or debate to assess outputs \cite{kumar-etal-2025-courteval, kim-etal-2024-debate}. By having agents play different roles (e.g. as domain experts, critics, defenders), the evaluation can incorporate diverse criteria and adversarial feedback, more closely emulating a panel of human judges.

Another motivation for agent-based judges comes from the rise of LLM-based agents – systems that use LLMs to plan, act, and solve complex tasks step-by-step. These agentic systems pose new evaluation challenges. Traditional evaluation often looks only at final task outcomes, ignoring the reasoning process, tool use, or intermediate steps that led there \cite{zhuge2024agentasajudgeevaluateagentsagents}. For autonomous agents that carry out extended sequences of actions (e.g. coding assistants \cite{zhang2024codeagentenhancingcodegeneration}, web navigation agents \cite{chae2025web}), focusing solely on success/failure at the end misses crucial insights into how and why the agent succeeded or failed. Human evaluation of each intermediate step is impractical at scale. This gap spurred the development of the Agent-as-a-Judge framework \cite{zhuge2024agentasajudgeevaluateagentsagents}, which explicitly uses an agent to evaluate another agent, examining the entire chain of actions and decisions rather than just the final answer. Agent-as-a-judge extends the LLM-as-judge idea by equipping the AI evaluator with agent-like capabilities – such as tool use, memory, and multi-step reasoning – so it can give rich intermediate feedback throughout the task-solving process. In other words, the judge is itself an autonomous agent that can observe and critique each step of another agent, providing a more fine-grained and informative evaluation.

In summary, the field has evolved from single-model judges toward more sophisticated multi-agent and agentic evaluation approaches to address the complexity of modern LLM applications. In the following, we review the prominent methods and frameworks for agent-as-a-judge and related multi-agent evaluation, compare their performance and design, and discuss applications across different domains. We highlight how these approaches aim to achieve the cost-effectiveness of LLM-based evaluation while improving alignment with the multi-dimensional perspectives of human judgment. We also identify open challenges and future research directions in leveraging LLM agents as reliable judges of each other’s outputs.

\section{Background and Motivation}
In this section, we introduce the main paradigms for LLM-based evaluation. We begin with single-model LLM-as-a-judge frameworks, then examine recent advances in multi-agent systems—such as debate and committee-based approaches—and conclude with the agent-as-a-judge paradigm, which enables step-by-step evaluation of autonomous agents. This progression sets the stage for understanding the motivations, benefits, and limitations that drive ongoing innovation in AI-based evaluation.
\subsection{LLM-as-a-Judge: Single Model Evaluators}
The LLM-as-a-judge paradigm uses a single LLM to evaluate text or agent behavior, often in a reference-free manner (without gold answers) \cite{li2024llmsasjudgescomprehensivesurveyllmbased}. Typically, a powerful model like GPT-4.1 is given a prompt describing evaluation criteria and one or more model-generated responses, and it outputs a score, ranking, or decision. There are three common modes: 
\begin{itemize}
\item Pointwise evaluation: The LLM judge scores one output at a time on given criteria (e.g. giving a coherence score) \cite{kim2024prometheus}.
\item Pairwise evaluation: The LLM judge compares two outputs and decides which is better \cite{saad-falcon-etal-2024-ares, cao2024compassjudger1allinonejudgemodel}. This is used in win-rate metrics, where the judge picks a winner between model A vs B for the same query.
\item Checklist evaluation: Checklist-based LLM evaluation frameworks break down complex assessment tasks into detailed, criteria-driven judgments to enable systematic and transparent analysis of model performance. Approaches like LLM-RUBRIC \cite{hashemi-etal-2024-llm} utilize multi-dimensional rubrics and personalized calibration networks, while CheckEval \cite{DBLP:journals/corr/abs-2403-18771} leverages LLMs to generate rubrics and automates part of the evaluation process using predefined taxonomies and seed questions. Complementing these, RAGCHECKER \cite{ru2024ragchecker} advances fine-grained evaluation for retrieval-augmented generation by extracting atomic factual claims from both model outputs and gold answers, then applying bidirectional entailment to assess precision and recall at the claim level. Collectively, these checklist-based methods offer a structured and granular way to evaluate LLMs across diverse scenarios.
\end{itemize}

LLM judges typically assess a range of qualities such as fluency, correctness, relevance, factuality, and adherence to instructions \cite{li2024llmsasjudgescomprehensivesurveyllmbased}. For example, a judge might be prompted to check if an answer is factually accurate and logically consistent, or whether it follows the user’s request and is well-written. Early works like G-Eval \cite{liu-etal-2023-g} demonstrated that an LLM can perform surprisingly well in evaluating open-ended responses, achieving strong correlation with human raters on tasks like dialogue response quality. In fact, automatic LLM-based benchmarks such as MT-Bench \cite{zheng2023judgingllmasajudgemtbenchchatbot} and AlpacaEval \cite{alpaca_eval} have reported high Spearman correlations (often 0.8–0.9) between GPT-4’s judgments and aggregate human preferences \cite{zheng2025cheating}. This high alignment has made LLM judges popular for rapid benchmarking of models in leaderboards (e.g. LMSYS’s Chatbot Arena uses GPT-4 judgments).

How do these LLM judges work under the hood? Often, they are prompted with detailed instructions and sometimes example evaluations (few-shot prompts \cite{brown2020languagemodelsfewshotlearners}) to guide their judging behavior. Some frameworks use chain-of-thought prompting \cite{wei2023chainofthoughtpromptingelicitsreasoning}, where the judge is asked to reason step-by-step about the output’s quality before giving a score. Despite these efforts, single LLM judges have known biases and vulnerabilities \cite{zheng2023judgingllmasajudgemtbenchchatbot}. They can be gamed by adversarial outputs: recent research showed that even a nonsense “null” response can trick a GPT-4 evaluator into giving high rankings if crafted in a persuasive style \cite{zheng2025cheating}. Models also tend to favor outputs that resemble their own phrasing or length – a bias towards verbosity and “familiar” styles \cite{dubois2024lengthcontrolled, chen-etal-2024-humans}. These limitations motivated enhancements to the LLM-as-judge approach, leading to multi-agent and ensemble-based evaluators.

\subsection{Multi-Agent Judges: Debate and Committee Frameworks}
Instead of relying on a single omniscient model, multi-agent approaches employ multiple LLMs that interact – via debate, discussion, or voting – to evaluate content. The intuition is that by having agents challenge each other’s opinions, the evaluation becomes more robust and closer to a multi-annotator human process \cite{kim-etal-2024-debate, kumar-etal-2025-courteval}. We outline several prominent frameworks in this category:

\begin{itemize}
\item ChatEval \cite{chan2024chateval} introduced a multi-agent referee team that debates the quality of a given response. Multiple LLM agents are assigned distinct personas (simulating different expertise or viewpoints) and engage in a group discussion about the strengths and weaknesses of the response. Through this collaborative debate – inspired by how humans brainstorm or discuss in panels – the agents come to a conclusion on the rating. An important finding from ChatEval is that diversity of agent roles is critical: if all agents are prompted with the same persona or perspective, the benefits diminish. Thus, one agent might focus on factual accuracy, another on linguistic style, another on user relevance, etc., mirroring a committee of judges with different priorities. ChatEval demonstrated superior accuracy and human-correlation on evaluating open-ended answers compared to single GPT-4 judges. In evaluations on open-ended question answering \cite{vicuna2023} and dialogue response generation \cite{gopalakrishnan2019topical} benchmarks, the multi-agent debate strategy improved correlation with human judgments by roughly 10–16\% over single-agent prompting. This highlights that synergy among agents can correct individual model biases and yield more reliable scores.

\item DEBATE \cite{kim-etal-2024-debate} is a framework explicitly adding an adversarial critic agent. It uses three agents: a Scorer (proposes an initial evaluation or score), a Critic (plays “devil’s advocate” to find faults in the Scorer’s judgment), and a Commander (coordinates the interaction). The Critic agent’s role is to surface biases or blind spots in the initial evaluation – for example, pointing out if the Scorer was too lenient on factual errors. The Commander helps iterate this debate and maintain context. Through this adversarial dialogue, the final evaluation is refined. DEBATE was shown to outperform prior state-of-the-art single-agent evaluators on meta-evaluation benchmarks like SummEval \cite{fabbri-etal-2021-summeval} (for summary quality) and TopicalChat  \cite{gopalakrishnan2019topical} (for dialogue). This indicates the devil’s advocate mechanism effectively improves judgment quality, addressing the “inherent limit” of a single LLM judge by resolving some of its biases. Notably, DEBATE found that the extent of debate (how many critique rounds) and the personas used can influence performance, suggesting a trade-off between thorough discussion and efficiency.
\item CourtEval \cite{kumar-etal-2025-courteval} takes inspiration from courtroom dynamics to structure the evaluation process. It defines three roles for LLM agents: a Grader (Judge), a Critic (Prosecutor), and a Defender (Defense Attorney). First, the Grader assigns an initial score to the model’s output. Then the Critic argues why this score might be incorrect or which aspects of the output are problematic (similar to a prosecutor challenging a claim). Next, the Defender argues in support of the original output, countering the Critic’s points. Finally, the Grader revises the score after considering both sides’ arguments, yielding a more balanced decision. This adversarial yet balance-seeking process is analogous to how multiple human judges or jurors might debate a verdict. Empirically, CourtEval showed substantial gains in evaluation quality, significantly outperforming previous methods on the SummEval \cite{fabbri-etal-2021-summeval} and TopicalChat benchmarks \cite{gopalakrishnan2019topical}. By explicitly modeling opposing perspectives, it closes much of the gap between automatic evaluation and human judgment quality. CourtEval’s design also reinforces the notion that multiple LLMs can improve one another through debate and cooperation – as observed in other studies where groups of models yield more factual and complete answers than a lone model \cite{10.5555/3692070.3692537, li2024prd}.
\item MAJ-EVAL \cite{chen2025multiagentasjudgealigningllmagentbasedautomated} stands for Multi-Agent-as-Judge Evaluation and emphasizes multi-dimensional human-like evaluation. This framework addresses two noted issues in prior LLM-as-judge work: (1) ad-hoc design of agent personas \cite{li-etal-2024-coevol}, and (2) lack of generality across tasks \cite{li2025llmgeneratedpersonapromise, 10.1145/3586183.3606763}. MAJ-EVAL introduces a systematic two-step solution. In Step 1, the system automatically constructs evaluator personas by extracting key dimensions from relevant domain documents. For example, given research papers or guidelines in a domain, it might derive that for medical text evaluation, important dimensions are “clinical accuracy” and “readability,” while for an educational QA task, dimensions like “grade-level appropriateness” and “engagement” matter. Each dimension is used to create a distinct persona for an agent (e.g., a “Clinician” agent focusing on accuracy, or a “Teacher” agent focusing on engagement). In Step 2, multiple agents grouped by stakeholder type engage in intra-group debates, and then an aggregator agent combines their feedback into final scores. Notably, agents are organized to reflect real stakeholder groups (e.g., in a healthcare summary task, agents representing doctors, patients, and caregivers each hold discussions, and their consensus is aggregated). This design simulates how diverse human experts would evaluate a system from different angles. MAJ-EVAL has been demonstrated in education and medical domains, where it generated evaluations better aligned with expert human ratings than both conventional metrics and single-agent judges. By automating persona creation and using debates per group, it improved scalability and adaptability of multi-agent evaluation, avoiding the need to hand-craft roles for each new task.
\end{itemize}

In essence, these multi-agent judge frameworks share a common theme: multiple LLMs acting in concert to evaluate, often with at least one agent playing a critical or adversarial role. This arrangement harnesses the power of “many eyes” on the problem, aiming to cancel out individual model idiosyncrasies. Empirically, approaches like ChatEval, DEBATE, CourtEval, and MAJ-EVAL have all reported stronger agreement with human judgments and more robust evaluations than single-model methods. The exact mechanisms differ – some use explicit debates and rebuttals, while others use parallel assessments and aggregation – but all yield a kind of ensemble judgment that benefits from model diversity.



\subsection{Agent-as-a-Judge: Evaluating Agents with Agents}

While the above methods largely focus on evaluating static outputs (e.g., answers to questions, summaries of text), the Agent-as-a-Judge \cite{zhuge2025agentasajudge} framework extends these ideas to evaluating dynamic agent behavior. Agent-as-a-Judge was motivated by the observation that LLM-based agents (which perform multi-step tasks using reasoning and tools) were not well-served by evaluations that only look at final outcomes. For example, if two AI coding agents attempt a programming task, a pure outcome metric (pass/fail) or even a final answer quality score might not reveal how each agent approached the task or where one excelled over the other in process. Agent-as-a-Judge proposes to use an agent to evaluate another agent, thereby enabling an evaluation of the entire trajectory rather than just the end result.

In practice, an “agent judge” is an autonomous LLM-based agent endowed with similar abilities as the agents it evaluates – it can observe intermediate steps, utilize tools (if needed), and perform reasoning over the agent’s action log. This judge agent often runs in parallel with the agents being evaluated, monitoring their decisions at each step of a task. Crucially, it can give granular feedback: not only a final score, but also pinpoint which requirements were satisfied or which steps were efficient/correct. In the original work, Agent-as-a-Judge was applied to evaluate AI code generation agents working on DevAI - a benchmark of 55 realistic automated AI development tasks (e.g., fixing a bug, implementing a feature) with hierarchical requirements \cite{zhuge2025agentasajudge}. The judge agent could check, for instance, whether the code compiled at intermediate stages, whether the agent followed each user requirement, and how many attempts or tool calls were used. This yielded a richer evaluation than just whether the final code worked.

The judge agent evaluates agent performance on each sub-requirement and the overall task, producing detailed scores. Notably, Agent-as-a-Judge preserved the cost-effectiveness of LLM-based evaluation while enhancing informativeness. It does not require a human in the loop for intermediate checking, yet its judgments were found to be as reliable as human evaluations in identifying the better agent solution. In fact, the study reported that Agent-as-a-Judge dramatically outperformed a standard LLM-as-a-judge that only saw final outputs, and achieved parity with human evaluators on the code tasks. For example, in one comparison, the agent judge’s decisions differed from the human-majority vote only ~0.3\% of the time, whereas a single LLM judge disagreed ~31\% of the time. Moreover, the agent judge even exceeded any individual human evaluator in consistency, aligning most closely with the collective human judgment. This suggests that an AI agent, by virtue of tirelessly analyzing step-by-step details, can nearly replace human evaluators for certain complex process-oriented tasks[90].

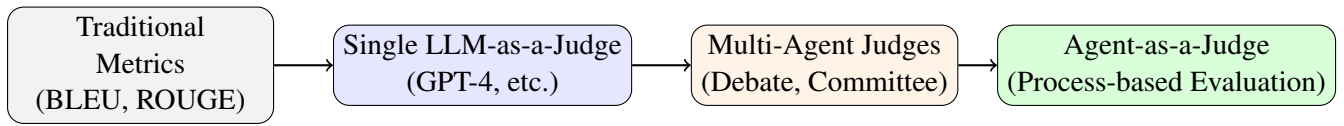
\begin{figure*}[ht]
\centering
\begin{tikzpicture}[node distance=1cm, every node/.style={align=center}, scale=1, transform shape]
  \node (metrics) [draw, rounded corners=6pt, fill=gray!10, minimum width=3.5cm, minimum height=1cm] {Traditional \\ Metrics \\ (BLEU, ROUGE)};
  \node (llmjudge) [draw, rounded corners=6pt, fill=blue!10, right of=metrics, xshift=3.5cm, minimum width=3.5cm, minimum height=1cm] {Single LLM-as-a-Judge \\ (GPT-4, etc.)};
  \node (multiagent) [draw, rounded corners=6pt, fill=orange!10, right of=llmjudge, xshift=3.5cm, minimum width=3.5cm, minimum height=1cm] {Multi-Agent Judges \\ (Debate, Committee)};
  \node (agentjudge) [draw, rounded corners=6pt, fill=green!15, right of=multiagent, xshift=3.5cm, minimum width=3.5cm, minimum height=1cm] {Agent-as-a-Judge \\ (Process-based Evaluation)};
  
  \draw[->, thick] (metrics) -- (llmjudge);
  \draw[->, thick] (llmjudge) -- (multiagent);
  \draw[->, thick] (multiagent) -- (agentjudge);
\end{tikzpicture}
\caption{Evolution of LLM Evaluation Paradigms: from traditional automatic metrics, to single-model LLM judges, to multi-agent debate systems, and finally to agent-as-a-judge frameworks that can evaluate processes and autonomous agents.}
\label{fig:evaluation-evolution}
\end{figure*}

In summary, Agent-as-a-Judge represents an “organic extension” of the LLM-as-judge idea into the realm of autonomous agents \cite{zhuge2025agentasajudge}. It is particularly useful when evaluating systems where the journey (reasoning trajectory) matters as much as the destination (final answer). By providing intermediate rewards or critiques, an agent judge can also facilitate training improvements, which we discuss later as a future direction. Figure~\ref{fig:evaluation-evolution} shows the conceptual evolution of evaluation frameworks for LLMs.

\section{Committee and Ensemble-Based Evaluators}

Another related thread in the literature is the use of multiple models in an ensemble to improve evaluation robustness. This is somewhat orthogonal to the debate frameworks – it is more about combining judgments from different models or runs. For instance, one might take multiple LLM-as-judge models (say GPT-4.1 and another strong model) and aggregate their scores, or use an ensemble voting among several prompts or few-shot examples to stabilize a single model’s outputs. The principle is that aggregating independent judgments can reduce variance and error, akin to a voting committee. Some multi-agent evaluators effectively do this with a structured communication beforehand (as in debates), while simpler ensemble methods might just average scores from multiple runs or models.

A striking example of leveraging multiple models’ strengths is the Mixture-of-Agents (MoA) approach \cite{wang2025mixtureofagents}. Although MoA was primarily proposed to enhance generation capability, it implicitly provides an evaluation boost by iteratively refining answers with multiple agents. In MoA, several LLMs (possibly of different types) are organized in layers; each agent in a new layer gets to see the outputs of agents from the previous layer and then generates an improved response. Through this multi-agent, multi-step refinement, MoA was able to produce very high-quality answers – so high that on automatic benchmarks judged by GPT-4 (like AlpacaEval and MT-Bench), the MoA’s final answers outranked even GPT-4’s own answers. In fact, an ensemble of open-source models in a MoA framework achieved a win rate of 65.1\% on AlpacaEval 2.0, substantially surpassing GPT-4’s 57.5\% on the same benchmark. This suggests that a committee of weaker models can outperform a single strong model when they effectively share information and vote on a solution. In evaluation terms, MoA demonstrates the “wisdom of crowds” effect: by exposing each agent to others’ outputs, the models correct each other’s mistakes, leading the judge to prefer the consensus result. This result supports the idea that diverse model ensembles can yield more robust outcomes, and by extension, an ensemble of judges might similarly improve evaluation stability (though MoA used the ensemble to improve generation, not explicitly as separate judges).

In practice, some benchmarks now incorporate multiple LLM judges \cite{tan2025judgebench, zhou2025evaluating} or multiple prompts \cite{efrat-etal-2023-lmentry, liu2024agentbench} as an ensemble to get more reliable evaluation scores. The aim is to average out idiosyncratic errors – for example, if one judge is overly harsh on length, another might counterbalance it. There are also toolkit-like frameworks (e.g., OpenAI Evals \cite{openai2023evals}) that allow using multi-annotator support, effectively running several LLM evaluations in parallel and aggregating. All these methods align with the core insight of agent-as-a-judge: employing multiple AI perspectives yields a richer and often fairer assessment than any single metric or model alone.

\section{Comparative Analysis of Evaluation Approaches}

Having outlined the key methods, we now compare them along several axes: reliability and human-alignment, granularity of feedback, cost and complexity, and applicability to different scenarios. Table \ref{tab:framework-comparison} summarizes major approaches, while the text below discusses their relative strengths and weaknesses:

\begin{table*}[ht]
\centering
\caption{Comparison of Major Agent-Based LLM Evaluation Approaches}
\begin{tabular}{|p{3.0cm}|p{3.2cm}|p{3.0cm}|p{3.5cm}|}
\hline
\textbf{Framework} & \textbf{Interaction Style} & \textbf{Key Strengths} & \textbf{Main Limitations} \\
\hline
LLM-as-a-Judge & Pointwise, pairwise, or checklist & Fast, scalable, transparent, good for general tasks & Susceptible to bias, style/preference lock-in, limited domain expertise \\
\hline
ChatEval & Multi-agent debate, diverse personas & Captures diverse perspectives, higher alignment with human judgments & Prompt/persona design complexity, computational cost \\
\hline
DEBATE & Adversarial discussion (Scorer, Critic, Commander) & Bias mitigation, robust evaluation, adversarial error detection & Increased complexity and cost, tuning critique depth \\
\hline
CourtEval & Courtroom-style adversarial evaluation (Judge, Critic, Defender) & Balanced, transparent, models real-world deliberation & Prompt and protocol complexity, cost \\
\hline
MAJ-EVAL & Persona-based group debates and aggregation & Stakeholder realism, automatable persona construction, high human-alignment & Role/dimension selection, scalability \\
\hline
Agent-as-a-Judge & Step-by-step process and action evaluation & Granular feedback, tracks process not just outcome, close to human-level for agent tasks & Computational overhead, reproducibility, stability \\
\hline
Human Panel & Manual annotation, panel/jury consensus & Gold standard for subtle and subjective tasks & Cost, time, inconsistency, not scalable \\
\hline
\end{tabular}
\label{tab:framework-comparison}
\end{table*}

\begin{itemize}
\item Human Evaluation (Baseline): Human judges remain the most trusted standard, especially for subjective qualities and for final validation. They can consider subtle context and ethical nuances. However, human evaluation is expensive, slow, and not reproducible at scale. Individual humans can be inconsistent or biased, and multi-annotator approaches are needed for reliable results. In dynamic agent tasks, humans also struggle to monitor every step without specialized expertise or tools. Thus, while indispensable for certain fine judgments, human eval doesn’t meet the rapid iteration needs of modern LLM development.

\item Automatic Metrics (BLEU/ROUGE/etc.): These provide objective, repeatable scores and work well for tasks with clear expected outputs (e.g., translation). But in open-ended generation, their low correlation with quality is well documented. They cannot capture semantic correctness or user satisfaction adequately. As such, they often serve as rough proxies or filters, but not as final say for high-stakes assessment.

\item Single LLM-as-a-Judge: This method strikes a balance by being much faster and cheaper than human eval, while handling nuanced language better than surface metrics. A well-calibrated LLM judge (especially GPT-4 class) shows impressively high agreement with aggregate human preferences (often in the 0.7–0.9 Spearman range) \cite{liu-etal-2023-g}. It also offers explanations for its scores, increasing transparency (e.g., highlighting a factual error it found). The limitation is that a single LLM judge can introduce systematic biases – for example, preferring outputs written in a style it was trained on, or being vulnerable to adversarial prompt tricks \cite{ zheng2025cheating}. Studies have found length biases (longer answers appearing better) and self-model bias (favoring models similar to the judge) in such evaluations \cite{zheng2023judgingllmasajudgemtbenchchatbot}. Moreover, if the judge model lacks domain knowledge (say evaluating a legal argument without legal training), it may miss important errors – thus single LLM judges can struggle in highly specialized domains without additional context.

\item Multi-Agent (Debate/Committee) Evaluators: By incorporating multiple agents and possibly adversarial roles, these frameworks generally achieve higher reliability and closer alignment to human consensus than a lone model. For instance, both CourtEval \cite{kumar-etal-2025-courteval} and DEBATE \cite{kim-etal-2024-debate} demonstrated new state-of-the-art correlations with human judgments, beating prior single-model methods by a significant margin. The advantage stems from diversity of perspectives: an error that one agent overlooks might be caught by another. The debate mechanism also forces justification of scores, leading to more well-founded decisions. However, this comes at a cost: multi-agent evaluations are computationally heavier – running three or more large models sequentially or in parallel for each evaluation. This can be 2–3x (or more) the cost of a single LLM judge per query, which adds up when benchmarking many examples. The complexity of prompt design also increases; one must carefully craft roles and interaction rules. There’s a risk that agents might converge to groupthink if not set up properly (e.g., if one dominant agent’s opinion sways others each time). Ensuring genuine disagreement and independent reasoning is key to avoiding all agents uniformly sharing the same blind spot. Some works address this by using different base models for different agents, but that can introduce fairness issues (as one model may unduly defer to another). Interestingly, Liang et al. (2024) found that an LLM judge tends to favor arguments made by an agent of the same model family, which can bias debate outcomes. They suggest using either the same model for all roles or completely different ones for each role to mitigate this bias. Despite these complexities, multi-agent evaluators are powerful in scenarios requiring multi-faceted analysis – their strength is in evaluating open-ended, complex tasks (creative writing, dialogs, etc.) where no single objective metric suffices.

\item Agent-as-a-Judge (Agentic Evaluators): These shine in evaluating process-oriented tasks and autonomous agent behaviors. The major strength is the granularity of evaluation: an agent-judge can give step-by-step feedback, identify which sub-goals failed, and even provide a narrative of an agent’s performance \cite{zhuge2025agentasajudge}. This is invaluable for diagnosing failures in complex agents (e.g., where in the chain of reasoning a mistake occurred). Agent judges can also interact with the same environment as the agent – for example, running a piece of code or querying a database – to independently verify the agent’s result. This goes beyond what a static LLM judge can do. The Agent-as-a-Judge framework was shown to match human evaluators in reliability on code tasks, and even outperform individual human judges in some cases by being more consistent. For researchers, this means agent judges could potentially serve as a surrogate for large panels of human testers, enabling rapid development of agentic systems. The downsides include computational cost (these judges often require running a full agent which might do web browsing, code execution, etc., incurring significant time per task). In Zhuge et al.’s experiments \cite{zhuge2025agentasajudge}, an agent-judge could take on the order of tens of minutes of compute per task in the worst case. Another consideration is stability: because agentic evaluations involve many moving parts (external tools, multi-step interactions), the results might have higher variance if any nondeterministic tool or API calls are involved. Ensuring reproducibility and fairness (the agent judge should treat all agents evaluated in the same manner) can be tricky. Nonetheless, for domains like software debugging, robotic planning, or any setting where the method is as important as the outcome, agent-based judges are uniquely suited.

\end{itemize}

In light of these comparisons, a hybrid evaluation strategy is often advisable. For example, pipeline an automatic LLM-based evaluation for broad filtering or initial scoring, and then employ a targeted multi-agent or agentic judge for deeper analysis of top contenders or specific complex cases. Human oversight can be reserved for edge cases or final confirmation. This layered approach can leverage the speed of LLM and the judgment of humans where each is best applied.

To illustrate comparative performance, consider quantitative benchmark results: On AlpacaEval \cite{alpaca_eval}, a single GPT-4 judge already achieves about 90\% agreement with crowd preferences, but methods like MoA that effectively ensemble multiple models pushed the evaluated win-rate higher (showing improved model outputs that the judge favored) \cite{wang2025mixtureofagents}. On TopicalChat \cite{gopalakrishnan2019topical}, ChatEval’s multi-agent discussion reached a Kendall Tau correlation ~0.57 with humans, versus ~0.52 by a single GPT-4 evaluator – a meaningful gain in a high range \cite{chan2024chateval}. And in the code generation domain, the agent judge framework obtained nearly perfect agreement with a majority vote of 5 human experts (used as ground truth), whereas a lone LLM judge lagged behind that level \cite{zhuge2025agentasajudge}. These figures demonstrate that while all LLM-based evaluators have made progress in approximating human judgment, those that incorporate multiple perspectives or follow the full process achieve the closest alignment.

In terms of scenarios of strength: single LLM judges are sufficient and very efficient for general-domain, moderately complex tasks (e.g., evaluating grammar, straightforward Q\&A responses). Multi-agent judges excel in highly subjective or multi-criteria tasks, like evaluating a story’s humor, coherence, and originality all at once – something a single metric cannot do. Agentic judges are unparalleled for interactive or long-horizon tasks (code writing, multi-turn tool use, game playing agents) where one needs to know not just who succeeded, but how. Each approach thus has its niche, and the choice depends on the evaluation objectives (outcome-focused vs. process-focused) and resource constraints.

\section{Domain-Specific Applications}
Agent-as-a-judge and multi-agent evaluation frameworks have been explored across various domains, often to address the particular needs of evaluating LLMs in specialized or sensitive contexts. Here we discuss applications in medicine, law, finance, and education, noting how each domain benefits from multi-agent or agent-based evaluation and any findings from those implementations.

\subsection{Medicine}
In the medical domain, accuracy and safety are paramount, and evaluations often require domain expertise that laypeople or general models lack. An example scenario is evaluating summaries of medical research or answers to health questions generated by LLMs \cite{deyoung-etal-2021-ms}. A single AI judge might miss subtle clinical inaccuracies, whereas a multi-agent approach can simulate a panel of medical experts. Chen et al. demonstrated this with MAJ-EVAL on a medical literature summarization task \cite{chen2025multiagentasjudgealigningllmagentbasedautomated}. They created agents with personas like “Medical Researcher”, “Clinician”, and “Public Health Consumer”, each focusing on different aspects such as evidence strength, clinical relevance, or layperson clarity. These agents debated and scored summaries, and the aggregated evaluation aligned more strongly with expert human ratings than standard metrics like ROUGE. For instance, MAJ-EVAL caught cases where a summary was fluently written but contained a hallucinated clinical claim – something ROUGE would not penalize if wording overlapped \cite{croxford2024evaluationlargelanguagemodels}, but a medically savvy agent would flag. The multi-agent judge thus ensured that criteria like “clinical consistency” and “evidence strength” were properly weighted in the evaluation.

Beyond summarization, multi-agent evaluation can aid in medical dialogue systems or diagnosis Q\&A. Different agents could represent, say, a doctor, a patient, and a medical ethicist to evaluate a chatbot’s response – checking factual correctness, patient readability, and appropriateness respectively. While specific published examples are sparse, the principle has been noted: medical AI evaluations need multiple stakeholder perspectives \cite{chen2025multiagentasjudgealigningllmagentbasedautomated}. Indeed, Yang et al. highlight that to evaluate patient summaries, one should integrate feedback from care providers, caregivers, and patients, as each has unique concerns \cite{yang2025recoverdesigninglargelanguage}. Agent judges can simulate these roles, as an alternative to convening real people from each group.

There are also safety-oriented judge frameworks in medicine. For example, MedEval \cite{he-etal-2023-medeval} might use an agent to critique responses for harmful recommendations. Conceptually one could have an agent-as-a-judge that attempts to detect if a medical answer would lead to dangerous actions, effectively evaluating an agent for safety compliance (similar to how R-Judge benchmarks risk awareness in agents) \cite{yuan-etal-2024-r}. This is critical in medicine because an incorrect but confident LLM answer could have severe consequences. An AI judge tuned to medical facts and protocols could serve as an intermediary safety check for LLM outputs before they reach end-users.

\subsection{Law}

Legal applications of LLMs (e.g., in drafting documents or answering legal questions) demand precision and often involve interpreting complex regulations. Evaluating legal text quality is challenging even for humans without legal training. One intriguing multi-agent approach in the legal domain is AgentsCourt \cite{he-etal-2024-agentscourt} which built a court debate simulation for legal decision-making agents. In AgentsCourt, multiple AI agents take on roles akin to a courtroom (prosecution, defense, judge) and debate a legal case or argument. The inclusion of legal knowledge bases helps them ground their arguments. While AgentsCourt is aimed at developing AI that can debate like lawyers, its evaluation mechanism effectively has a judge agent assessing which side made a better case and what the likely decision should be. This is analogous to agent-as-a-judge in that the judge agent must evaluate arguments in a legal context, requiring deep domain understanding.

Another work, CourtEval (discussed earlier), though used for general text evaluation, explicitly borrowed the courtroom metaphor \cite{kumar-etal-2025-courteval}. This highlights that the adversarial, multi-agent approach is quite suitable for law – after all, legal judgments in real life are often made by weighing opposing arguments. An AI judge in law could, for example, evaluate a contract summary by having one agent argue that the summary misses certain clauses and another agent defend the summary’s adequacy, with the judge resolving which argument is more convincing.

There is also the possibility of using LLM judges to evaluate AI-generated legal advice for correctness and ethical compliance. For instance, if an LLM gives tax advice, a panel of AI “jurors” might check it against tax code references. Some early works have created legal benchmarks for LLMs \cite{guha2023legalbench, goebel2024coliee_overview} (like checking case outcomes or statute reasoning), but incorporating agent judges is a potential next step. AgentsCourt suggests agent judges have been prototyped to decide outcomes of simulated legal cases. This could be extended to evaluating how well an AI followed legal reasoning: one agent could deliberately inject a fallacious argument and another agent (judge) must determine if the reasoning held is legally sound or not.
In summary, the law domain benefits from agent-as-a-judge by enabling simulated judicial processes. This not only checks correctness but also the rationale behind an AI’s legal answer. It aligns with the principle that in law, how you arrive at an answer (citing precedent, following logic) is as important as the answer itself.

\subsection{Finance}
Finance-focused LLM applications, such as financial report summarization, trading advice generation, or regulatory compliance checking, present unique evaluation criteria like numerical accuracy, risk awareness, and compliance with laws (e.g., avoiding insider information). Multi-agent judges have begun to appear here as well. FinCon \cite{yu2024fincon} applies the agent-as-a-judge paradigm by designing a multi-agent system inspired by real-world investment firm structures, where each agent (modeled using LLMs) has a distinct role—such as analyst agents specializing in various data modalities (text, audio, tabular data) and a manager agent who acts as the final decision-maker. Analyst agents independently process and judge streams of financial information (e.g., news, earnings calls, quantitative indicators), distilling them into actionable insights. The manager agent then evaluates (“judges”) these diverse inputs, synthesizes them, and makes the final trading or portfolio management decisions. 

Moreover, financial text often has a quantitative nature. So an LLM judge must be good at catching arithmetic or logical errors – something single GPT-4 judges can do to an extent, but a specialized agent with a calculator tool could do better. For instance, evaluating a budget report summary: a tool-using agent judge could recompute totals to ensure no calculation mistakes. This is a simple form of agent-as-judge where the judge agent verifies numerical consistency that humans or base LLMs might miss. Given the high stakes (financial errors can cost money or legal consequences), using an ensemble of AI judges – some checking math, some checking compliance – can greatly enhance trustworthiness of evaluations in finance.

\subsection{Education}

Educational applications involve LLMs generating content like tutoring responses, quiz questions, or essay feedback. Evaluating educational content requires considering the audience’s perspective (age, background) and pedagogical value, not just factual accuracy. Multi-agent judges have been explicitly applied here. MAJ-EVAL \cite{chen2025multiagentasjudgealigningllmagentbasedautomated} is applied for a children’s reading comprehension Q\&A generation task \cite{chen-etal-2024-storysparkqa}. In this case, they derived personas from education research: e.g., a “Teacher” agent focusing on correctness and curriculum fit, a “Parent” agent focusing on age-appropriate language and engagement, and even a “Child” persona evaluating if the question would be interesting or understandable to a young reader. These agents debated the quality of questions generated by a model for a children’s story, scoring aspects like Educational Appropriateness and Context Relevance. The results showed that agents with the relevant educational roles achieved higher correlation with human (teacher/parent) judgments than a one-dimensional automated metric. For example, an agent judging as a teacher was very aligned with human teachers on identifying if a question was appropriate for a 5-year-old’s comprehension level. This underscores the value of multi-agent evaluation in education: it’s able to capture qualitative, developmental factors that generic metrics would ignore.

In an educational context, diversity of perspectives is again crucial. A single LLM judge might give high marks to a technically correct explanation, but a student might find it confusing. A multi-agent setup can reflect this by having a “student” agent judge clarity and a “tutor” agent judge completeness, for instance. Using the same role for all agents is detrimental \cite{chan2024chateval} – if every agent were a “strict grammarian”, they might all fixate on grammar and miss whether the content is engaging or conceptually correct. The ChatEval study noted that distinct personas (e.g., one focusing on creativity, another on coherence) improved performance. Education content evaluation benefits from this because one must balance grammar, factual accuracy, engagement, and age suitability.

In summary, in education the agent-as-a-judge approach allows capturing the holistic educational value of content – something inherently multi-faceted (is it correct, clear, engaging, level-appropriate?). Early results in children’s QA tasks are promising, showing higher agreement with what human educators value \cite{chen2025multiagentasjudgealigningllmagentbasedautomated}. We can expect similar approaches to be valuable in other educational AI evaluations, such as assessing AI-generated tutoring feedback or personalized learning plans, where multiple perspectives (student, teacher, pedagogue) matter.

\section{Limitations of Agent-Based Evaluation}
Despite their promise, current agent-as-a-judge and multi-agent evaluation methods face several limitations and open challenges:
\begin{itemize}
\item Bias and Impartiality: Ironically, while multi-agent frameworks were created to reduce bias, they can introduce new biases. As noted, if the judge agent shares the same model or training biases as one of the participants, it may favor that side \cite{chan2024chateval}. It is observed that an LLM judge “consistently favors the side with the same LLM backbone” in a debate, which is problematic for fairness \cite{liang-etal-2024-encouraging}. They suggest using homogeneous roles (all agents use the same base model) or completely heterogeneous models to avoid one biasing toward another. Ensuring impartiality remains challenging, especially if all agents come from the same pre-trained model (they might share blind spots and biases). Diverse model pools or calibrated objectives for each agent might mitigate this.

\item Collusion or Mode Collapse: In multi-agent debates without proper adversarial setup, agents might converge too quickly or implicitly collude. For instance, if all agents are clones of a cooperative dialogue model, they might agree with each other politely rather than truly debate. This yields little benefit over a single agent. Designing prompts to encourage genuine disagreement (Devil’s advocate roles \cite{kim-etal-2024-debate}, explicit instructions to find counterpoints) is necessary. There is a fine line, however, because too adversarial a setup can lead to agents nitpicking trivial issues or getting stuck in argumentative loops, which doesn’t reflect constructive evaluation. Tuning the “tit for tat” aggressiveness is non-trivial. Liang et al. discuss how the degree of opposition in debate (and an adaptive mechanism to stop debating when enough consensus is reached) affects outcomes. Improper tuning could cause either degeneration (no new ideas after a point) or endless debates \cite{liang-etal-2024-encouraging}.

\item Cost and Scalability: Running multiple large models is expensive. A single GPT-4 call might be acceptable, but CourtEval’s judge+critic+defender triples the calls \cite{kumar-etal-2025-courteval}. Agent-as-a-judge might involve an agent planning and executing code, which could take minutes per instance \cite{zhuge2024agentasajudgeevaluateagentsagents}. This can be a bottleneck if one wants to evaluate thousands of examples. Researchers are looking into optimizing this, e.g., by using smaller but specialized models for some roles \cite{belcak2025smalllanguagemodelsfuture}. One interesting finding is that using a weaker judge with strong debaters can still yield good results, whereas a strong judge with weak debaters is worse \cite{liang-etal-2024-encouraging}. This implies that using a cheaper (slightly weaker) model for the judge role, while investing in stronger agents that produce arguments, might be a cost-effective trade-off. Nonetheless, at present, multi-agent eval is mainly used in research settings where quality is prioritized over speed. For deployment or continuous evaluation, the compute cost needs to be managed.

\item Domain Expertise Limitations: While we can assign an agent the persona of a “doctor” or “lawyer”, the agent only has as much actual expertise as its training data and prompt can induce. If the underlying model isn’t truly knowledgeable in that domain, its evaluation will suffer. For example, an LLM might not reliably detect a subtle medical contraindication in an answer because it doesn’t actually have medical training, just a surface-level persona. Chen et al. addressed this by feeding actual domain literature into the persona construction \cite{chen2025multiagentasjudgealigningllmagentbasedautomated}, but not all domains have easily available corpora or clearly defined dimensions. In highly specialized or novel domains, AI judges might lack the necessary knowledge to judge correctly, leading to false assurances. Therefore, calibration with human experts is still needed – either via incorporating expert-written evaluation guidelines or by having humans review some agent-judgments for critical errors.

\item Evaluation of the Evaluators (Meta-evaluation): A recurring issue is: how do we know the agent-judges are correct? The community relies on meta-evaluation benchmarks (like measuring correlation with human judgments on shared datasets) \cite{kumar-etal-2025-courteval, kim-etal-2024-debate}. While improved correlation is encouraging, it is not a perfect measure of true reliability. For instance, high correlation could mean the AI judge is good, but it could also mean humans had biases that the AI simply learned to mimic. There have been cases where LLM judges agree with average human scores but might fail on edge cases where humans with deeper thought would disagree \cite{posner2025llm_judicial}. Also, an AI judge might overfit to the training signals – if it was indirectly tuned to match human dataset scores, it might not generalize to novel kinds of content. This is analogous to overfitting a metric to a benchmark. To truly validate agent-judges, we would need continued human oversight and spot-checks, especially when extending to new tasks.

\item Cheatability and Robustness: Zheng et al. demonstrated that current LLM evaluators can be fooled by cleverly crafted outputs that exploit their evaluation prompts \cite{zheng2025cheating}. Multi-agent setups could be more robust, since a deceptive output would have to fool not one but multiple agents (and possibly a debate where one agent might call out the deception). However, multi-agent judges could have their own failure modes. If all agents are from the same model family, an exploit that confuses that model could trick all of them (e.g., a weirdly formatted answer that throws off their parsing). Moreover, a malicious actor could try to design outputs that specifically exploit the known rules of a framework like CourtEval – for instance, phrasing an answer to preemptively counter criticisms (thus fooling the Critic agent). As agent-based eval becomes more standard, robustness against adversarial responses will be vital. Techniques from adversarial ML and red-teaming will likely be applied to test these evaluators. Initial steps include controlling for output length and style biases, but more dynamic defenses (like adversarial training of the judge agents) could be developed \cite{alon2023detectinglanguagemodelattacks, Xie2023DefendingCA, robey2023smoothllm}.

\item Complexity of Implementation: From a practical standpoint, setting up these multi-agent systems is more complex than using a single metric. Researchers must design prompts for each agent, figure out how agents pass information (turn-taking protocol), and decide how to aggregate outputs (is it a majority vote, an average score, or does the judge have the final say?). There isn’t yet a one-size template. For broader adoption, there will need to be easier tools or libraries that encapsulate these patterns (similar to how today one can call a BLEU function easily). Until then, there’s a barrier to entry.

\item Limited Scope of Current Benchmarks: Many of the results showing success for agent-as-a-judge are on benchmarks specifically chosen by the method’s authors (e.g., DevAI for agentic coding tasks, SummEval for NLG tasks). We have fewer assessments of how these methods perform on completely different tasks or in the wild. There is a need to test generalizability: Does a framework like MAJ-EVAL work equally well for, say, evaluating an AI-generated financial report or an interactive game agent? Possibly new unseen dimensions or interactions would challenge it.

\end{itemize}

In light of these limitations, researchers approach agent-judges as augmented evaluators rather than infallible oracles. Human oversight remains important, especially in high-stakes use (medicine, legal). The goal is not to eliminate humans, but to reduce their burden and catch obvious issues automatically. Many current efforts combine automated and human evaluation – for instance, using AI judges to pre-score and filter outputs, and then having humans review a smaller set or the cases where AI judges disagreed. See Figure~\ref{fig:agent-eval-limitations} for a checklist of key limitations discussed in this section.

\begin{figure*}[ht]
\centering
\begin{tikzpicture}
\filldraw[fill=blue!6, draw=blue!40, thick, rounded corners=10pt] (0,0) rectangle (13,7.5);

\filldraw[fill=blue!20, draw=none, rounded corners=7pt] (0,7) rectangle (13,7.5);
\node[anchor=west, font=\bfseries\large, text=blue!70!black] at (0.4,7.28) {Checklist of Limitations in Agent-Based Evaluation};

\node[anchor=west, font=\Large, text=green!60!black] at (0.8,6.5) {\checkmark};
\node[anchor=west, font=\normalsize] at (1.5,6.5) {Potential bias and impartiality (model and group bias)};

\node[anchor=west, font=\Large, text=green!60!black] at (0.8,5.7) {\checkmark};
\node[anchor=west, font=\normalsize] at (1.5,5.7) {Collusion or ``mode collapse'' among agents};

\node[anchor=west, font=\Large, text=green!60!black] at (0.8,4.9) {\checkmark};
\node[anchor=west, font=\normalsize] at (1.5,4.9) {High computational cost and limited scalability};

\node[anchor=west, font=\Large, text=green!60!black] at (0.8,4.1) {\checkmark};
\node[anchor=west, font=\normalsize] at (1.5,4.1) {Limited true domain expertise (surface-level personas)};

\node[anchor=west, font=\Large, text=green!60!black] at (0.8,3.3) {\checkmark};
\node[anchor=west, font=\normalsize] at (1.5,3.3) {Challenges in meta-evaluation (how do we evaluate the evaluators?)};

\node[anchor=west, font=\Large, text=green!60!black] at (0.8,2.5) {\checkmark};
\node[anchor=west, font=\normalsize] at (1.5,2.5) {Susceptibility to adversarial manipulation and cheatability};

\node[anchor=west, font=\Large, text=green!60!black] at (0.8,1.7) {\checkmark};
\node[anchor=west, font=\normalsize] at (1.5,1.7) {Implementation complexity and prompt engineering overhead};

\node[anchor=west, font=\Large, text=green!60!black] at (0.8,0.9) {\checkmark};
\node[anchor=west, font=\normalsize] at (1.5,0.9) {Limited scope of current benchmarks (generalizability concerns)};
\end{tikzpicture}
\caption{Key limitations of agent-based evaluation frameworks.}
\label{fig:agent-eval-limitations}
\end{figure*}

\section{Future Directions}
Agent-as-a-judge and multi-agent evaluation methods are rapidly evolving. Based on the current state and literature, we identify several key avenues for future research and improvements:
\begin{itemize}

\item Expanding to New Domains and Tasks: Current works have piloted agent judges in a few areas. Future research should test these frameworks on a wider array of domains, especially those with unique challenges. Some examples: creative writing (poetry or story generation), coding assistant agents (beyond just code correctness, evaluating code quality and documentation), conversational agents providing emotional support (where sensitivity and psychological nuance matter). Each new domain might require tweaking persona design or debate formats. A robust agent-as-a-judge framework should be generalizable, able to incorporate new dimensions without a ground-up redesign. This may involve building libraries of common evaluation dimensions (fluency, factuality, etc.) and modular persona templates that can be composed for a domain. Additionally, multimodal LLMs (which handle images, audio, etc.) are emerging – future agent-judges might need to evaluate outputs that are not just text, e.g., an image caption or a generated graph. Multi-agent evaluation in a multimodal setting (imagine one agent checking if an image is relevant to a caption, another checking caption grammar) is largely unexplored.

\item Data and Benchmark Development: To advance agent-based evaluation, we need more standard meta-evaluation benchmarks and datasets. These would allow comparing different evaluation strategies objectively (similar to how GLUE or SuperGLUE benchmark model capabilities). For example, a benchmark could contain a set of model outputs for various tasks with careful human scores along multiple axes. Researchers could then run their agent judge on this and see how well it reproduces each axis and the overall ranking. Some efforts in this direction include SummEval \cite{fabbri-etal-2021-summeval} (for summary quality), TopicalChat  \cite{gopalakrishnan2019topical} (for dialogue) and DevAI \cite{zhuge2025agentasajudge} (for code generation). However, more diverse and large-scale meta-evaluation sets, especially covering multi-turn interactions and real user queries, would encourage robust development. Also, collecting multi-stakeholder human evaluations (e.g., doctors + patients, teachers + students) will be valuable so that agent judges can be validated on multi-dimensional alignment, not just single-score correlation.

\item Reducing the Dependence on Proprietary LLMs: Currently, many evaluation frameworks lean on GPT-4.1 or similar top-tier models. An interesting direction is exploring whether smaller or open-source models (possibly specialized or fine-tuned for evaluation) can replace these, making evaluation more accessible. For instance, SambaNova’s research suggests a fine-tuned 40B model could replicate GPT-4’s judge decisions on AlpacaEval and MT-Bench \cite{raju2024llama405b_judge}. Developing dedicated evaluation models – potentially using instruction tuning on lots of comparison data – could yield judges that are efficient and on par with the best LLMs. This ties with the idea of Mixture-of-Experts for judges \cite{xu2024perfectblendredefiningrlhf}, where different sub-models in an ensemble handle different criteria. By distributing the load, one could use smaller models for specific checks (like a toxicity classifier agent) alongside a main model for general judgment.

\item Integrating Tool Use in Judges: Agent judges that can use tools (like search engines, code interpreters, calculators) will become more important as tasks get complex or require live data verification. We might see hybrid evaluator agents: for example, an agent that, when evaluating a factual answer, automatically consults a knowledge base or the web to fact-check statements. This would dramatically improve evaluations of factual correctness and reduce hallucination going unnoticed. Tool-using judges could also simulate user interaction – e.g., if evaluating a dialogue agent that queries a database, the judge agent could attempt the same query to see if the answers match. Some initial work in agent-as-a-judge already hints at this (the code evaluation agent that runs code \cite{zhuge2024agentasajudgeevaluateagentsagents}). Generalizing it, one could incorporate modules for different tools and let the judge decide when to deploy them.

\item Self-Improvement and Training Feedback: One of the most exciting prospects of agent-as-a-judge is using it to train better models without human labels. Zhuge et al. allude to a “flywheel effect” where evaluated agents and judge agents iteratively improve each other \cite{zhuge2024agentasajudgeevaluateagentsagents}. For example, an agent judge gives intermediate feedback on an agent’s reasoning; that feedback can be used as a reward signal in reinforcement learning or to prompt the agent to reflect and revise. This becomes a form of self-play, akin to how AlphaGo self-play improved Go playing \cite{silver2016alphaGo} – here, an agent judge and an agent performer engage in a loop that sharpens both. Future research might formalize this: training an agent entirely through signals from an agent judge (which itself might be improved through occasional human calibration). If successful, this could massively scale learning – imagine an AI writing answers and another AI grading them with high fidelity to human standards, so we can generate essentially unlimited training pairs. There are risks (they might reinforce each other’s biases), but techniques like periodically resetting or introducing human-written checks could control that. The concept is similar to Debate for AI safety \cite{irving2018aisafetydebate} where two AIs debate and the outcome (as judged by a human) trains them to be truthful. A fully automated variant might use an agent judge instead of a human, once the judge is deemed reliable. Achieving that is a long-term goal with substantial research needed in ensuring the judge’s values align with ours.

\item Addressing Robustness and Adversarial Exploits: As discussed in limitations, future agent-judge systems must be robust. Research into adversarial evaluation will likely expand – e.g., creating challenging test cases designed to fool the evaluators and then improving them. One direction is to use one agent to generate tricky outputs and another to judge, in a red-team/blue-team setup, and iteratively train the judge on these hard cases (an adversarial training loop). Another idea is enabling judges to indicate uncertainty or flag cases where they are not confident (perhaps because the debate was inconclusive). Rather than forcing a possibly wrong judgment, an AI judge could say, “I’m not certain; escalate to human.” Designing uncertainty estimates or triggers for fallback to human evaluation would make these systems safer in practice.

\item User Experience and Dynamic Evaluation: In some settings, evaluation is not just offline but part of an interactive system. For instance, a tutoring agent might assess its own answer before showing it to a student, in real-time. This dynamic use of agent-as-a-judge (like an inner monologue check) could improve the live performance of AI agents. Research can explore how fast and efficiently an AI agent can judge itself on the fly – perhaps by training a smaller internal critic model that runs alongside the main model (a form of internal multi-agent debate inside one model’s forward pass). This bleeds into model architecture innovations: maybe future LLMs will architecturally include a “judge module” that continuously evaluates the outputs of a “generator module.”

\item Ethical and Transparency Considerations: With AI judges taking roles that affect model deployment (e.g., gating content), it’s crucial to ensure they are transparent and their value judgments are appropriate. Future work should consider auditing AI evaluators – making sure, for example, a content moderation judge agent isn’t systematically biased against certain dialects or perspectives. Tools to interpret why an AI judge gave a score (beyond just a text explanation, perhaps mapping to known criteria or highlighting portions of text that influenced the decision) will increase trust. Another aspect is calibrating strictness: in education, an overly harsh AI grader could discourage students; in law, an overly lenient evaluator could miss harmful content. Tuning the judges to appropriate “rubrics” likely requires human-in-the-loop at development time.
\end{itemize}

In conclusion, the future of agent-as-a-judge research is rich with possibilities. The overarching trend will be making these AI evaluators more reliable, more general, and more integrated into the AI development cycle. Evaluation should become an ongoing, holistic part of the agent development pipeline (“Evaluation-driven Development”) \cite{xia2025evaluationdrivendevelopmentllmagents}. Agent judges, possibly monitoring agents continuously in deployment and feeding back to developers, could be a realization of that vision – essentially AIs that help us oversee other AIs at scale. Achieving that in a trustworthy way is a grand challenge that will require advances in both algorithms and how we validate them against human standards.
\section{Conclusion}

The advent of agent-as-a-judge frameworks marks a significant evolution in how we evaluate AI systems. Moving beyond static metrics and limited human spot-checks, researchers are leveraging the very capabilities of LLMs – reasoning, perspective-taking, debating – to create dynamic evaluators that can judge AI agent outputs with greater depth and adaptability. We have reviewed how this paradigm emerged from the basic LLM-as-a-judge approach, addressing its shortcomings by incorporating multiple agents and agentic processes. Agent-as-a-judge, in its various incarnations (from debate teams to tool-using critic agents), offers a way to approximate the multi-faceted evaluation that complex tasks demand: it is cost-effective and scalable like automated metrics, yet closer to human-like appraisal in its richness of feedback.

We discussed concrete frameworks – ChatEval’s collaborative discussions, DEBATE’s devil’s advocate, CourtEval’s courtroom roles, MAJ-EVAL’s stakeholder alignment, and the meta-agent approach of Agent-as-a-Judge itself – and found that all share a common insight. No single evaluator (human or AI) sees the full picture; a chorus of voices leads to a fairer verdict. Empirically, multi-agent evaluators have raised the state of the art in alignment with human judgments on several benchmarks, and agent-judges have shown they can reliably evaluate even complicated multi-step tasks on par with expert humans. This suggests that these methods are not just academic curiosities, but practical tools for the AI community.

In domain-specific contexts like healthcare, law, finance, and education, we saw that agent-based evaluation is not only beneficial but perhaps necessary. These domains involve specialized criteria and high stakes, and multi-agent judges allow evaluation to be informed by domain knowledge and ethical perspectives without always requiring a human expert in the loop. For example, simulating a panel of medical experts or legal counsels via LLM personas can flag issues that generic evaluations would miss. The versatility of agent-as-a-judge means it can be tailored – you can dial in the types of “judges” you need for the task at hand, whether it’s a strict grammarian, a compassionate user perspective, a risk-averse regulator, or all of them together.

Despite these advances, we underscored that agent-based evaluation is not a solved problem. Challenges around bias, cost, consistency, and domain limits persist. An AI judge is only as good as the wisdom it has been given or has learned; thus, careful design and ongoing validation remain crucial. The community is actively working on these, and the trajectory is clear: future AI evaluation frameworks will likely feature hybrid teams of AI judges, possibly overseen by occasional human auditing – a structure akin to a human courtroom or committee, but operating at machine speed and scale.

In closing, agent-as-a-judge embodies a form of meta-intelligence: it’s intelligence evaluating intelligence. This meta level is important for creating AI that we can trust and continually improve. By letting agents serve as judges, we introduce feedback loops that can drive self-improvement (an AI learning from critique to refine itself). It opens the door to AI systems that are not just performers but also reflective critics. In the long run, such systems might contribute to safer AI – echoing the idea of “AI safety via debate” – by ensuring that for every answer an AI gives, another AI (or group of AIs) has scrutinized it from all angles for truthfulness and quality.

The literature reviewed suggests that while human evaluators will continue to play a key role, especially in setting the goals and standards, much of the routine heavy lifting of evaluation can be offloaded to these sophisticated AI judges. This collaboration between human and AI evaluation will enable faster development cycles and more robust AI deployments. As benchmarks grow and models further improve, we expect agent-as-a-judge strategies to become standard practice in AI evaluation – an essential component in the toolkit for governing and improving our increasingly powerful AI systems.

\end{document}